# Conditional Expressions for Blind Deconvolution: Multi-point form

S. Aogaki, I. Moritani, T. Sugai, F. Takeutchi, and F.M. Toyama, *Kyoto Sangyo University, Kyoto-603, Japan*

*Abstract*—We present conditional expression (CE) for finding blurs convolved in given images. The CE is given in terms of the zero-values of the blurs evaluated at multi-point. The CE can detect multiple blur all at once. We illustrate the multiple blur-detection by using a test image.

*Index Terms* Deconvolution, Image restoration.

## I. INTRODUCTION

Lane and Bates' (LB) blind deconvolution [1] is based on the zero-sheets of the *z*-transform of given images. This method enables us to uniquely eliminate blurs convolved in the given images. However, due to its advanced analytical method, the computational complexity for the image processing is enormous.

In our first paper [2] we presented the version 1 of conditional expressions (CEs) for LB's blind deconvolution. The CEs can detect the zero-values of the blurs automatically. The CEs of the version 1 are given in terms of the derivatives of the zero-values of the blurs. Hence, they can be evaluated at a single point. This is a great advantage of the version 1. On the other hand, for blurs of large sizes the CEs of the version 1 become very complex.

In this paper, we present a simple version of the CEs, which we call the version 2 throughout this paper. The CE of the version 2 is given in terms of only zero-values of blurs evaluated at multi-point. The version 2 reduces the computational complexity considerably compared with those of the version 1. Instead we need to optimize a sampling parameter for the multi-point. This version 2 can also detect blurs of any sizes all at once.

In Sec. II we present the CE of the version 2. In Sec. III we give an illustration of the multiple blur detection by using a test image. Section IV is for the summary.

## II. CONDITIONAL EXPRESSIONS FOR BLURS

We consider a situation where any noise is absent. In the situation a given (observed) image $g(x,y)$ can be modeled as the convolution of a true image $f(x,y)$ and a blur image $h(x,y)$, i.e.,

$$g(x,y) = f(x,y) * h(x,y), \qquad (1)$$

where the blur $h(x,y)$ is assumed to be caused by a linear shift system. Throughout this paper the sizes of the given image and the blur image are denoted by $M \times N$ and $m \times n$, respectively. The *z*-transform $H(u,v)$ of a $m \times n$ blur $h(x,y)$ is written as

$$H(u,v) = \frac{1}{mn} \sum_{x=0}^{m-1} \sum_{y=0}^{n-1} h(x,y) u^x v^y, \qquad (2)$$

where $u$ and $v$ are complex variables. In he following we consider zero-values $\beta_i(u)$ ($i=1,2,\cdots,n': n' \leq n-1$) which are the solutions of $H(u,v)=0$ for a given $u$.

The CE is derived as follows. From (2) we obtain the following simultaneous equations for $h(x,y)$ with the $\beta_i$ at $mn$ different points of $u_\ell$ ($\ell = 0, 1, \cdots, mn-1$),

$$\begin{aligned} H(u_0, \beta_i(u_0)) &= 0, \\ H(u_1, \beta_i(u_1)) &= 0, \\ &\vdots \\ H(u_{mn-1}, \beta_i(u_{mn-1})) &= 0. \end{aligned} \qquad (3)$$

From (3), the $mn \times mn$ matrix $D$ composed of the coefficients of the blur elements is given as

$$D = \begin{pmatrix} 1 & \beta_i(u_0) & \cdots & \beta_i(u_0)^{n-1} & u_0 & u_0\beta_i(u_0) & \cdots & u_0\beta_i(u_0)^{n-1} & \cdots\cdots & u_0^{m-1} & u_0^{m-1}\beta_i(u_0) & \cdots & u_0^{m-1}\beta_i(u_0)^{n-1} \\ 1 & \beta_i(u_1) & \cdots & \beta_i(u_1)^{n-1} & u_1 & u_1\beta_i(u_1) & \cdots & u_1\beta_i(u_1)^{n-1} & \cdots\cdots & u_1^{m-1} & u_1^{m-1}\beta_i(u_1) & \cdots & u_1^{m-1}\beta_i(u_1)^{n-1} \\ \vdots & \vdots & & \vdots & \vdots & \vdots & & \vdots & & \vdots & \vdots & & \vdots \\ 1 & \beta_i(u_q) & \cdots & \beta_i(u_q)^{n-1} & u_q & u_q\beta_i(u_q) & \cdots & u_q\beta_i(u_q)^{n-1} & \cdots\cdots & u_q^{m-1} & u_q^{m-1}\beta_i(u_q) & \cdots & u_q^{m-1}\beta_i(u_q)^{n-1} \end{pmatrix}, \qquad (4)$$

where $q = mn-1$ and we dropped the constant factor $1/mn$. The CE is given as $E^u_{m \times n}(\beta_i(u_0),\cdots,\beta_i(u_{mn-1})) \equiv |D| = 0$. This is actually the condition for obtaining non-trivial solutions for $h(x,y)$. The $E^u_{m \times n}(\beta_i(u_0),\cdots,\beta_i(u_{mn-1}))$ implicitly includes the CEs for any blurs of the sizes smaller than $m \times n$, except for the size $r \times 1$ ($r = 1,\cdots, m$). This can be proven by basic manipulations for the determinant. Further, the CE can detect

S. Aogaki is with the Department of Information and Communication Sciences, (tel.: +81-75-705-1694, e-mail: aogaki@cc.kyoto-su.ac.jp).
I. Moritani is with the Department of Information and Communication Sciences, (tel.: +81-75-705-1694, e-mail: i654168@cc.kyoto-su.ac.jp).
T. Sugai is with the Department of Information and Communication Sciences, (tel.: +81-75-705-1694, e-mail: sugai@cc.kyoto-su.ac.jp).
F. Takeutchi is with the Department of Computer Sciences, (tel.: +81-75-705-1694, e-mail: takeut@ksuvx0.kyoto-su.ac.jp).
F.M. Toyama is with the Department of Information and Communication Sciences, (tel.: +81-75-705-1898, e-mail: toyama@cc.kyoto-su.ac.jp).

the zero-values of blurs of the size $1 \times k$ $(k = n+1, \cdots, N)$. This can be seen as follows. In such a one dimensional blur the zero-values are all constant, i.e., do not depend on $u$. Hence when we substitute the zero-values of $1 \times k$ blurs into $|D|$, more than two row vectors become proportional to each other. This is why the CE constructed for the $m \times n$ blurs detects the zero-values of $1 \times k$ blurs.

This version 2 seems to be simple formally compared with the version 1 in the sense that we do not have to evaluate derivatives of the zero-values. However, there is a nervous situation in this version 2. In order to evaluate the CE, we have to solve $\beta_i$ at $mn$ different points of $u_\ell$ $(\ell = 0, \cdots, mn-1)$ numerically. Further, the $mn$ solutions must be taken to be of the same $\beta_i$. Further, when we reconstruct an original image by the inverse Fourier transform after removing the zero-values $\beta_i$ of the blurs from those of the given image, we have to find $\beta_i$ of the blurs at every step $u_j$ $(j = 0, \cdots, M-1)$. Hence, in each $u_j$, we have to solve $mn$ solutions of $\beta_i$. To do this we take $mn$ different points $u_\ell$ $(\ell = 0, 2, \cdots, mn-1)$ in the vicinity of each $u_j$, i.e., as $u_{j,\ell} = u_j + \ell \Delta u$. This requires an optimization for the parameter $\Delta u$. Namely, to hold the same $\beta_i$ we have to take $\Delta u$ considerably small. On the other hand, if we take $\Delta u$ too small, $\beta_i$ solved for $mn$ different points $u_{j,\ell}$ become close to each other. This may cause more than two row vectors in the determinant $|D|$ to be nearly proportional to each other. Thus, undesired $|D| \approx 0$ may be caused. Therefore, we have to determine an optimal $\Delta u$ to accomplish the detection process successfully.

In order to complete the image restoration we need the CE also for the variable $u$. The CE for $u$ is given in the similar form to (4). The CE for $u$ is denoted as $E_{m \times n}^v (\gamma_i(v_0), \cdots, \gamma_i(v_{mn-1})) = 0$, where $\gamma_i$ are the zero-values of $u$ for the blurs.

## III. ILLUSTRATION

In the preceding section we have given the version 2 of the CEs. In this section, we illustrate the multiple blur detection by using the same image as that used for the illustration of the version 1 [2].

Fig.1 shows the test images that are the same as those used in [2]. Fig. 1(a) shows a $40 \times 40$ model image that we regard as a true image [3]. Figs. 1(b), 1(c), 1(d), and 1(e) represent blur images of the sizes $1 \times 2$, $2 \times 1$, $2 \times 2$, and $2 \times 3$, respectively. We convolved these four blurs into the true image of Fig. 1(a). Fig. 1(f) shows the convolved image, of which size is $43 \times 44$.

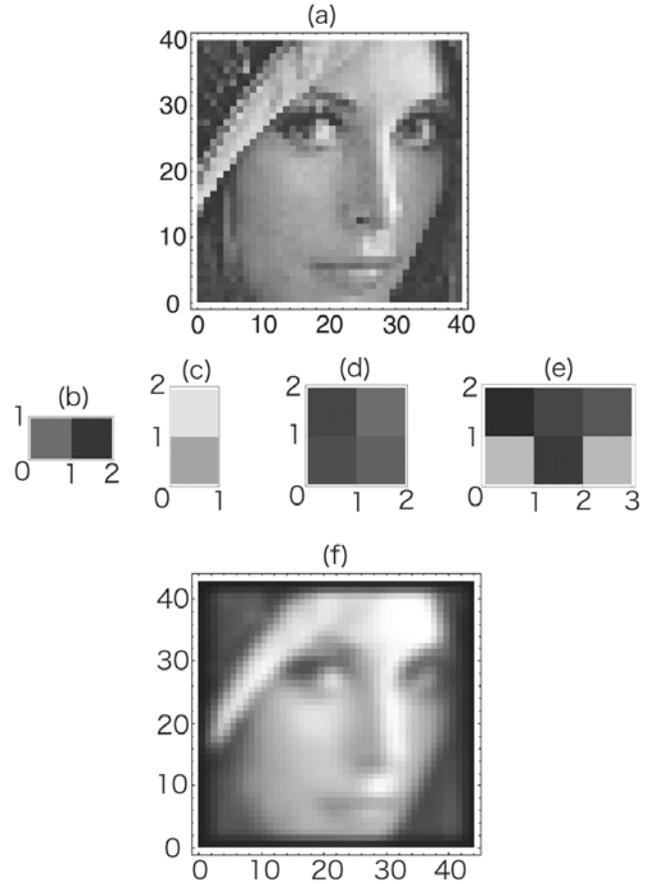

Fig. 1. (a): True image of $40 \times 40$ size that we took from [3]. (b): Blur image of $1 \times 2$ size. (c): Blur image of $2 \times 1$ size. (d): Blur image of $2 \times 2$ size. (e): Blur image of $2 \times 3$ size. (f): The image that was obtained by convolving the four blurs of (b)-(e) into the true image of (a). The size of the convolved image is $43 \times 44$.

In Fig. 2 we show the results of the numerical evaluations of the CEs $E_{2 \times 3}^u (\beta_i(u_0), \cdots, \beta_i(u_5))$ and $E_{2 \times 3}^v (\gamma_i(u_0), \cdots, \gamma_i(u_5))$, where we plotted $\log[|E_{2 \times 3}^u (\beta_i(u_0), \cdots, \beta_i(u_5))| \times 10^{50} + 1]$ and $\log[|E_{2 \times 3}^v (\gamma_i(v_0), \cdots, \gamma_i(v_5))| \times 10^{50} + 1]$. We optimized $\Delta u$ and $\Delta v$ as $\Delta u = \Delta v = \rho e^{-i \Delta \phi}$ with $\Delta \phi = \pi / 2150$ and took as $\rho = 1$. Note that we took $\Delta \phi$ very small to hold the same $\beta_i$ in sampling $mn$ points. As we mentioned earlier, this causes not true $E_{2 \times 3}^u (\beta_i(u_0), \cdots, \beta_i(u_5)) \approx 0$. To distinguish the true $E_{2 \times 3}^u (\beta_i(u_0), \cdots, \beta_i(u_5)) = 0$ from the not true $E_{2 \times 3}^u (\beta_i(u_0), \cdots, \beta_i(u_5)) \approx 0$ we need a high precision. This is why we multiplied the factor $10^{50}$ in evaluating the CE. As we stressed in the preceding section, both $E_{2 \times 3}^u (\beta_i(u_0), \cdots, \beta_i(u_5))$ and $E_{2 \times 3}^v (\gamma_i(u_0), \cdots, \gamma_i(u_5))$ include CEs for blurs of smaller sizes $1 \times 2$, $2 \times 1$, $2 \times 2$, and $2 \times 3$ implicitly. Therefore, $E_{2 \times 3}^u (\beta_i(u_0), \cdots, \beta_i(u_5))$ must detect totally four $(=1+1+2)$ zero-values $\beta_i$. When there exists a degenerate zero-value in the $2 \times 3$ blur, it may be three $(=1+1+1)$. On the other hand, the number of zero-values $\gamma_i$ that should be detected by $E^v_{2\times 3}(\gamma_i(u_0),\cdots,\gamma_i(u_5))$ is just three $(=1+1+1)$. As seen in Fig. 2(a), for $\phi_u=0$, four zero-values $\beta_1$, $\beta_{13}$, $\beta_{20}$, and $\beta_{21}$ are detected by $E^u_{2\times 3}(\beta_i(u_0),\cdots,\beta_i(u_5))$. Also at other $\phi_u$ four zero-values are well detected by $E^u_{2\times 3}(\beta_i(u_0),\cdots,\beta_i(u_5))$. Here, note that the detected solution numbers of the zero-values are different at each $\phi_u$. In this illustration we numerically solved the zero-values by *Mathematica*, separately at each $\phi_u$. Accordingly, the order of the numerical solutions (zero-values) is at random at each $\phi_u$. As seen in Fig. 2(b), for $\phi_v=0$, three zero-values $\gamma_1$, $\gamma_3$, and $\gamma_{12}$ are detected by $E^v_{2\times 3}(\gamma_i(u_0),\cdots,\gamma_i(u_5))$. Also at other points $\phi_v$ three zero-values are well detected by $E^v_{2\times 3}(\gamma_i(u_0),\cdots,\gamma_i(u_5))$. Note that the detected zero-values by $E^u_{2\times 3}(\beta_i(u_0),\cdots,\beta_i(u_5))$ and $E^v_{2\times 3}(\gamma_i(u_0),\cdots,\gamma_i(u_5))$ are perfectly the same as those detected by the CEs of the version 1 [2]. This is quite natural because we used the same image for the two illustrations. This implies that the detection processes have been done successfully for both the two versions of the CEs.

Fig. 3 shows the restored image by removing the blurs. The restored image is perfectly the same as the original image of Fig. 1(a).

The version 2 of the CE is formally simpler than the version 1 of [2] because the version 2 of the CE can be evaluated with only the zero-values. However, we have to spend a time to optimize the sampling parameters $\Delta u$ and $\Delta v$.

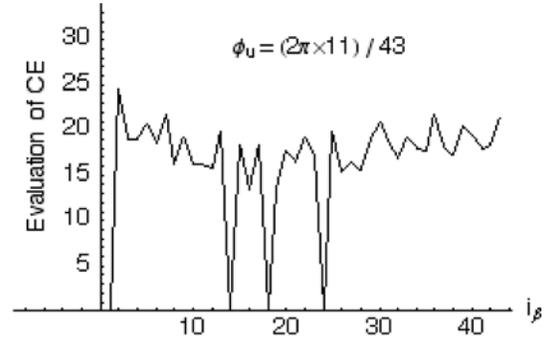

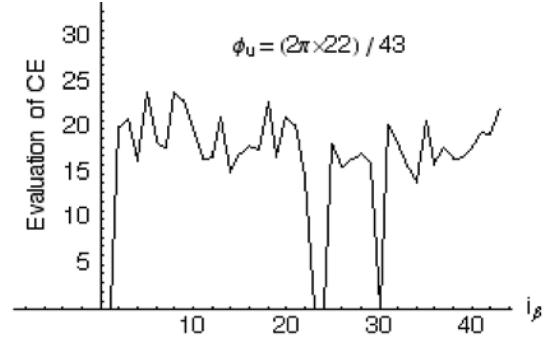

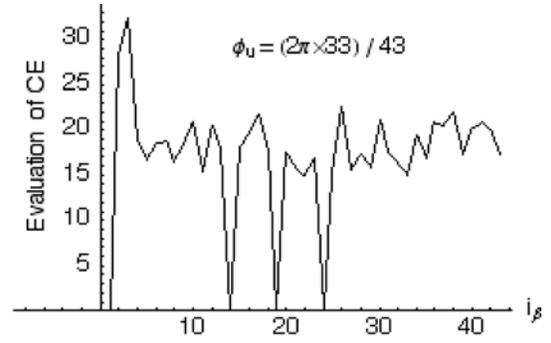

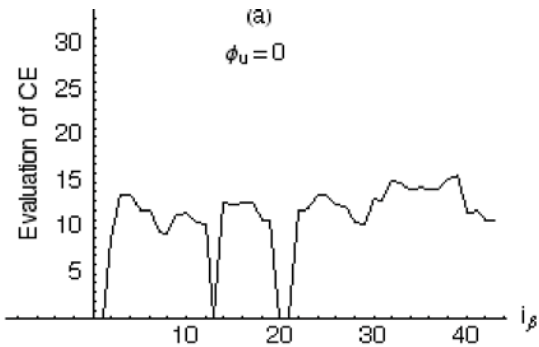

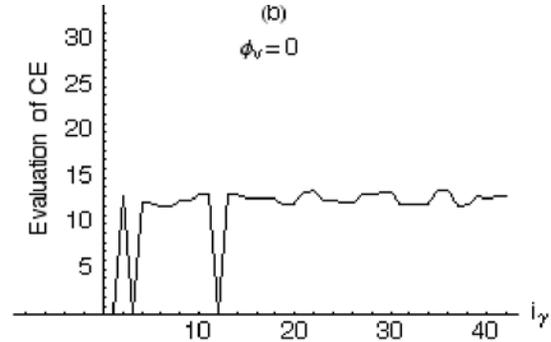

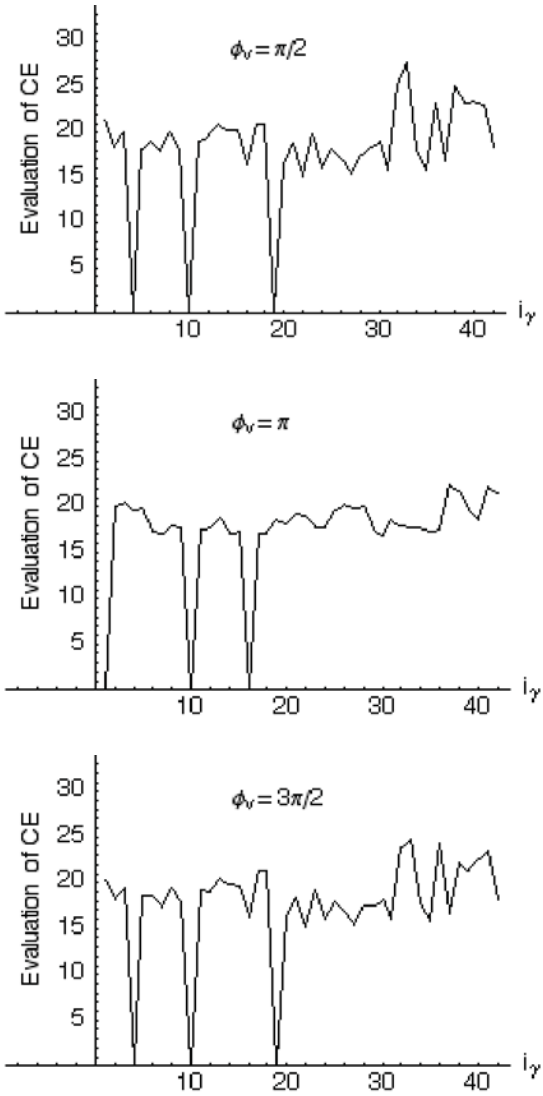

**Fig. 2.** Results of the evaluations of the version 2 of the CEs $E^u_{2\times 3}(\beta_i(u_0),\cdots,\beta_i(u_5))$ and $E^v_{2\times 3}(\gamma_i(v_0),\cdots,\gamma_i(v_5))$. (a) and (b) show the results of the tests done for $\beta_i$ and $\gamma_i$, respectively. We took $\rho_u$ and $\rho_v$ as $\rho_u = \rho_v = 1$.

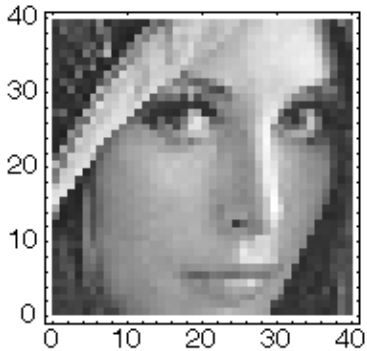

**Fig. 3.** Restored image by removing the four and three zero-values that were respectively detected by $E^u_{2\times 3}(\beta_i(u_0),\cdots,\beta_i(u_5))$ and $E^v_{2\times 3}(\gamma_i(v_0),\cdots,\gamma_i(v_5))$.

## IV. Summary

We have derived the version 2 of the CEs for finding multiple blur convolved in a given image, in a situation where any noise is absent. The CE is given in terms of the zero-values of the given image evaluated at multi-point. By using the CE we can avoid a tough analysis of the zero-sheets of the given image and reduce the processing time in the LB's deconvolution. The CE constructed for $m \times n$ blurs can actually detect any blurs of any sizes smaller than $m \times n$ all at once. Therefore, when we apply the CE for an actual image, it is desirable to construct the CE for blurs of as large size as possible.

As we mentioned above the CE detects multiple blur all at once. Instead the sizes of the detected blurs cannot be determined, except for a single blur. In our third paper [4] we present a simple method to find a single blur of only a specified size $m \times n$.

### Acknowledgement

This work was supported in part by the Science Research Promotion from the Promotion and Mutual Aid Corporation for private Schools of Japan, and a grant from Institute for Comprehensive Research, Kyoto Sangyo University.